\documentclass[10pt, a4paper]{article}

\usepackage{lrec-coling2024} 
\usepackage{amssymb}
\usepackage{multirow} 
\usepackage{txfonts}

\newcommand\blfootnote[1]{%
  \begingroup
  \renewcommand\thefootnote{}\footnote{#1}%
  \addtocounter{footnote}{-1}%
  \endgroup
}
\title{EthioMT: Parallel Corpus for Low-resource Ethiopian Languages}

\name{\normalsize Atnafu Lambebo Tonja $^{\spadesuit,\Diamondblack, \ast}$, Olga Kolesnikova $^{\spadesuit}$,  \\
\textbf{\normalsize Alexander Gelbukh $^{\spadesuit}$, Jugal Kalita $^{\clubsuit}$,} \\\\
\footnotesize
\address{
 $^\spadesuit$ Instituto Politécnico Nacional, Mexico, $^\Diamondblack$ Lelapa AI, 
\\
 \footnotesize
 $^\clubsuit$ University of Colorado Colorado Springs, USA
}} 
\abstract{
Recent research in natural language processing (NLP) has achieved impressive performance in tasks such as machine translation (MT), news classification, and question-answering in high-resource languages. However, the performance of MT leaves much to be desired for low-resource languages. This is due to the smaller size of available parallel corpora in these languages, if such corpora are available at all. NLP in Ethiopian languages suffers from the same issues due to the unavailability of publicly accessible datasets for NLP tasks, including MT. To help the research community and foster research for Ethiopian languages, we introduce EthioMT -- a new parallel corpus for 15 languages. We also create a new benchmark by collecting a dataset for better-researched languages in Ethiopia. We evaluate the newly collected corpus and the benchmark dataset for 23 Ethiopian languages using transformer and fine-tuning approaches. 
 \\ \newline \Keywords{Parallel corpus, EthioMT, Machine Translation, low resource language, Ethiopian languages} 
}

\begin{document}

\maketitleabstract

\blfootnote{$^\ast$ Work done during an internship at the University of Colorado Colorado Springs.}
\section{Introduction}
In recent years, due to advances in deep learning approaches such as the development of transformers \cite{vaswani2017attention}, machine translation (MT), a core task in natural language processing (NLP), has shown dramatic improvements in terms of coverage and translation quality \cite{wang2021progress}. It is well-known that a critical requirement for advancing MT is the availability of parallel corpora. The availability of parallel corpora is also necessary to facilitate the incorporation of languages in MT applications like Google Translation, Bing, and DeepL \cite{van2019translation}. The majority of the languages in the world do not have access to such translation tools since only a few high-resource languages have received significant attention \cite{tonja2023natural}. 

Most models and methods developed for high-resource languages do not work well in low-resource settings \cite{costa2022no,tonja2023natural,king2015practical}. Low-resource languages have also suffered from language technology designs \cite{joshi2019unsung,tonja2022improving}. Creating powerful novel methods for language applications is challenging when resources are limited and only a small amount of even unlabeled data is available. The problem is exacerbated when no parallel dataset exists for specific languages \cite{joshi2020state,ranathunga2023neural,adebara-abdul-mageed-2022-towards}.

Ethiopia is a country that stands out for its remarkable cultural and linguistic diversity, with over 85 spoken languages \cite{woldemariam2007challenges}. Only a few languages of Ethiopia have received attention in the area of NLP research and application development.  Most languages have been left behind due to resource limitation \cite{costa2022no,tonja2023natural}. It is hard to find publicly available datasets for Ethiopian languages to pursue NLP research because many researchers do not make their datasets publicly accessible \cite{tonja2023natural}. The unavailability of benchmark datasets and results for NLP tasks, including MT, makes research for newcomers and interested parties very difficult. This is obviously more difficult for languages with limited data in different digital forms. 

This paper introduces EthioMT: a parallel corpus for low-resource Ethiopian languages paired with English, and a benchmark dataset and experimental results for 23 Ethiopian languages. Our contributions are the following:
\noindent \textbf{(1)} We create a \textbf{new parallel corpus for 15 Ethiopian languages} paired with English.
   \noindent \textbf{(2)}  We introduce the \textbf{first benchmark dataset and results for relatively better resourced Ethiopian} (Amharic, Afaan Oromo, Tigrinya and Somali) \textbf{languages}. 
    \noindent \textbf{(3)} We evaluate MT performance with the \textbf{new corpus and present benchmark results}.
    \noindent \textbf{(4)} We \textbf{open-source} the parallel corpus to foster collaboration and facilitate research and development in low-resource Ethiopian languages.

\section{Related work} \label{related_work}
\textbf{Ethiopian languages} are categorized as low-resource due to the unavailability of resources for NLP tasks, including MT \cite{tonja2023natural}. Although MT is a better-researched area for Ethiopian languages compared to other NLP applications \cite{tonja2023natural}, only a handful of languages have received adequate attention from researchers.  \\
\textbf{Researched Languages} Compared to other Ethiopian languages, the following languages have received significant attention from researchers. Nevertheless, the collected corpora are not found in one location. It is hard to find benchmark datasets in these languages and datasets and associated results to reproduce and compare MT approaches. \\
\textit{Amharic} - Researchers have collected parallel datasets and proposed different MT approaches for Amharic-English translation \cite{kenny2018machine,teshome2012preliminary,hadgu2020evaluating,ashengo2021context,biadgligne2022offline,belay2022effect,gezmu2021neural,gezmu2021extended,biadgligne2021parallel}. \\
\textit{Afaan Oromo} - Similarly, there have been attempts to create Afaan Oromo-English MT datasets \cite{meshesha2018english, solomon2017optimal,adugna2010english,chala2021crowdsourcing, gemechu2021machine}. \\
\textit{Tigrinya} - For Tigrinya-English MT, researchers have attempted to create parallel datasets \cite{tedla2016effect,tedla2017morphological,berihu2020enhancing,azath2020statistical,kidane2021exploration}.\\
\textbf{Multilingual MT} Some researchers have included Ethiopian languages with other languages in multilingual MT systems.  \citet{lakew2020low} collected and created benchmark results for five African languages, including those mentioned above from Ethiopia.  \citet{costa2022no}, \citet{goyal2022flores} and \citet{fan2021beyond} included Ethiopian languages in their multilingual MT models and benchmark test sets. \citet{vegi2022webcrawl} crawled a multilingual parallel dataset for African languages, including Amharic and Afaan Oromo from Ethiopia.\\
\textbf{Other languages} There have been efforts to create and collect MT datasets for other Ethiopian languages. For example, \citet{tonja2021parallel} presented a parallel corpus for four low-resourced Ethiopian languages (Wolaita, Gamo, Gofa, and Dawuro).

\begin{table*}[!ht]
  \resizebox{\textwidth}{!}{
\begin{tabular}{llllll}
\hline
\multicolumn{1}{l|}{Language} &
  \multicolumn{1}{l|}{Family} &
  \multicolumn{1}{l|}{Explored prev.} &
  \multicolumn{1}{l|}{\begin{tabular}[c]{@{}l@{}}No. of \\ Speaker\end{tabular}} &
  \multicolumn{1}{l|}{Domain} &
  \multicolumn{1}{l}{Size} \\ \hline
 Afar (aar)&Afro-Asiatic / Cushitic  &$\times$ &1.5M  &Religious  &11K  \\
Afaan Oromo (orm) &Afro-Asiatic / Cushitic  & \checkmark  & 37M  &Misc  &\textbf{\underline{2.9M}} \\
 Awngi (awn)& Afro-Asiatic / Cushitic &$\times$&490K  &Religious  &7K  \\
 Amharic (amh)& Afro-Asiatic / Ethio-Semitic  &  \checkmark &57M  &Misc  & \textbf{\underline{1.5M}} \\
 Basketo (bst)& Afro-Asiatic/ Omotic & $\times$  &93K  &Religious  & 7K\\
 Dawuro (dwr)& Afro-Asiatic/ Omotic & \checkmark &1.5M  &Religious  &7K  \\
 Dashenech (dsh)& Afro-Asiatic/ Cushitic &$\times$ &99K  &Religious  &7K  \\
 Geez (gez)&Afro-Asiatic / Ethio-Semitic  & $\times$ &UNK  &Religious  &7K  \\
 Gamo (gmv)&Afro-Asiatic / Omotic  &\checkmark&1.09M   &Religious  &7K \\
 Gofa (gof)&Afro-Asiatic / Omotic   &\checkmark  & 392K &Religious  &7K  \\
 Gurage (sgw)&Afro-Asiatic / Ethio-Semitic  &$\times$  & 5.8M &Religious  &28K  \\
 Hadiya (hdy)& Afro-Asiatic / Cushitic &$\times$   & 1.3M &Religious  &28K  \\
 Kafa (kbr)& Afro-Asiatic / Omotic &$\times$ &830K  &Religious  &28K \\
Korate (kxc)& Afro-Asiatic / Cushitic  &$\times$   & 500K  &Religious  & 7K \\
 Majang (mpe)&Nilo-Saharan / Eastern Sudanic  &$\times$  &66K  &Religious  &9K  \\
 Male (mdy)&Afro-Asiatic / Omotic  &$\times$  &105K  &Religious  &7K  \\
 Murule (mur)&Nilo-Saharan / Eastern Sudanic  &$\times$   &300K  &Religious  &9K \\
 Nuer (nus)&Nilo-Saharan /Eastern Sudanic  &$\times$   &900K  &Religious  & 29K \\
 Shakicho (moy)&Afro-Asiatic / Omotic  &$\times$ &80K  &Religious  & 7K \\
 Sidama (sid)&Afro-Asiatic / Cushitic  &$\times$  &4M  & Religious &28K  \\
 Somali (som) &Afro-Asiatic / Cushitic  &\checkmark &22.3M  &Misc &\textbf{\underline{1.2M}}\\
 Tigrinya (tir) &Afro-Asiatic / Ethio-Semitic  & \checkmark  & 9M &Misc  &\textbf{\underline{140K}} \\
Wolaytta (wal)&Afro-Asiatic / Omotic   &\checkmark  &7M  &Religious  &29K  \\\hline
\end{tabular}
}
\caption{Languages and dataset details for \textbf{EthioMT} corpus. It shows languages, language families, the number of speakers, the domain, and the size of the collected dataset. In domain column \textbf{Misc} indicates \textit{mixed} corpus collected from religious, news, and other sources.   \textbf{\underline{Bold and underlined}} size indicates a dataset collected from different repositories and published works and merged into one dataset for the language to create a benchmark dataset}
\label{tab:language}
\end{table*}

\section{EthioMT}
\subsection{Discussion of Languages}
In this section, we enumerate languages included in the EthioMT corpus. Languages include in the EthioMT corpus belong to Afro-Asiatic and Nilo-Saharan language families. 
\subsubsection{Afro-Asiatic language family}
The Afro-Asiatic language family comprises about 250 languages spoken in North Africa, parts of sub-Saharan Africa, and the Middle East. Languages belonging to this family are grouped into six sub-groups: Berber, Chadic, Cushitic, Egyptian, Omotic, and Semitic \cite{epstein1998language}. EthioMT contains languages belonging to the Omotic, Cushitic, and Semitic sub-groups.

 \noindent\textbf{1) Omotic Languages} are a group of languages spoken in southwestern Ethiopia, in the Omo River region. The Ge'ez script is used to write some of the Omotic languages and the Latin script for others \cite{amha2017omotic}. Languages belonging to this group that we included in EthioMT are given below. 

\textbf{\textit{Basketo}} is spoken in the Basketo special woreda of the South Ethiopia Regional State. The Basketo language is also called Basketto, Baskatta, Mesketo, Misketto, and Basketo-Dokka. The speakers call the language "Masketo", while their neighbors call it "Basketo". The language has two dialects, Doko (Dokko) and Dollo (Dollo).

\textbf{\textit{Dawuro}} is a language spoken by about 1.09 million people in the Dawro zone of the South West Ethiopia Peoples' Region. It is also known as Dauro, Dawragna, Dawrogna, Ometay, Cullo, or Kullo. The language has four dialects: Konta, Kucha, Longkhai, and Yawngkon. 

\textbf{\textit{Gamo}} is spoken by around 1.63 million people in the Gamo Zone of the South Ethiopia Regional State. The speakers call the language Gamotstso. 

\textbf{\textit{Gofa}} refers to the language spoken in the Gofa zone of the South Ethiopia Regional State with around 392,000 speakers. 

 \textbf{\textit{Kafa}}, also known as Kefa or Kafi noono is a North Omotic language spoken in Ethiopia. It is spoken by about 830,000 people in the Keffa Zone in the South West Ethiopia Peoples' Region. The language is mainly spoken in and around the town of Bonga. 
 
 \textbf{\textit{Male}} is spoken in the Omo Region of Ethiopia. The Male people maintain their language vigorously despite exposure to outside pressures and languages.  
 
 \textbf{\textit{Shakicho}}, also known as Mocha, Shakacho, or Shekka, is spoken in the Sheka Zone of southwestern Ethiopia. It is closely related to Kafa. Loan words from Majang and Amharic influence the language's vocabulary. 
 
\textbf{\textit{Wolaytta}} is a North Omotic language spoken by the Welayta people in the Wolayita Zone of Ethiopia. It is estimated that 2 million people speak Wolaytta.

\noindent\textbf{2) Cushitic languages} are spoken primarily in the Horn of Africa, including Djibouti, Eritrea, Ethiopia, Somalia, and Kenya \cite{comrie2002languages}. The Cushitic languages use the Latin and Ge'ez script. Languages belonging to this family that are included in the EthioMT group are discussed below. 

\textbf{\textit{Afar}} is spoken by the Afar people in Ethiopia, Eritrea, and Djibouti. It is also known as Afar Af, Afaraf, and Qafar af. 
 About 1.5 million people speak Afar, the closest relative to the Saho language. 
 
\textbf{\textit{Afaan Oromo}}, also known as Oromo, is spoken by about 37 million people in Ethiopia, Kenya, Somalia, and Egypt. It is the third-largest language in Africa and the largest language in the Cushitic group in terms of speakers. The Oromo people are the largest ethnic group in Ethiopia and account for more than 40 percent of the population. 

\textbf{\textit{Awngi}} is a Central Cushitic language spoken by about 400,000 people in northwestern Ethiopia. 
 It is also known as Awiya, Awi, Agaw, Agau, Agew, Agow, Awawar, and Damot. 
 Most speakers live in the Agew Awi Zone of the Amhara Region. 
 Awngi is an Afro-Asiatic language spoken in parts of the Metekel Zone of the Benishangul-Gumuz Region. 
 
\textbf{\textit{Dashenech}} is also known as Dasenech, Daasanech, or Daasanach. The Daasanach people speak it in Ethiopia, South Sudan, and Kenya. The Daasanach people primarily live in the Lower Omo Valley of southwestern Ethiopia, along the eastern shore of Lake Turkana in Kenya, and in some parts of South Sudan. 

\textbf{\textit{Hadiya}} is spoken by the Hadiya people of Ethiopia. The language is also known as Hadiyyisa, Hadiyigna, Adiya, Adea, Adiye, Hadia, Hadiya, and Hadya. It is a Highland East Cushitic language. 
The Hadiya people are an ancient indigenous group in the southern part of Ethiopia. There are 1.4 million speakers of the Hadiya language, with 1.25 million of them speaking it as their mother tongue.  

\textbf{\textit{Korate}} is a Lowland East Cushitic language spoken by the Konso people in southwest Ethiopia. It has approximately 500,000 native speakers. The language has five dialects: Duuro, Fasha, Karatti, Kholme, and Komso. 
 The two main dialects are Fasha and Karatti. 
Konso is closely related to Dirasha (also known as Gidole). It is used as a "trade language" or lingua franca beyond the area of the Konso people. 
The Konso people are a Cushitic ethnic group who live in large towns in south-central Ethiopia. 

\textbf{\textit{Sidama}}, or Sidaamu Afoo, is a Cushitic language spoken by the Sidama people in southern Ethiopia. It uses the Latin alphabet. Almost nine million people speak Sidama. 
 It is the official language of the Sidama National Regional State (SNRS) and is used as a medium of instruction in primary schools.  Sidama is a branch of the Highland East Cushitic family.  
 
\textbf{\textit{Somali}} is the official language of Somalia, spoken by 6.5 million people. It is also spoken in Ethiopia, Djibouti, and Kenya. The total number of speakers worldwide is estimated at nearly 22 million. Its closest relative is the Oromo language, spoken in parts of Ethiopia and Kenya. Other related languages include Afar and Saho.

\noindent\textbf{3) Semitic languages} belong to a subfamily of the Afro-Asiatic language family, including Hebrew, Aramaic, Arabic, and Ethiopic. Most scripts used to write Semitic languages are abjad. Abjad refers to an alphabetic script that omits some or all vowels. Languages belonging to this group that we study are given below.
 
\textbf{\textit{Amharic}} is spoken by the Amhara and other regions in Ethiopia. 
 It is the second most-spoken Semitic language in the world, after Arabic. 
 Amharic is the official language of Ethiopia and has been since the 14th century. 
 It is also spoken in other countries, including Eritrea, Canada, the United States, and Sweden. 
Amharic is written using graphemes called \textit{fidal}, which means "script", "alphabet", "letter", or "character". 

\textbf{\textit{Ge'ez}} is an ancient Semitic language that originated in Eritrea and northern Ethiopia. 
 Ge'ez is believed to be around 5,000 years old, making it older than Hebrew and other Northern Semitic languages. 
Orthodox and Catholic churches in Eritrea and Ethiopia still use it as a liturgical language. 
Ge'ez went extinct as a natural language over 1,000 years ago. It was written in two systems: an abjad and later an abugida. 

\textbf{\textit{Gurage}} is spoken by the Gurage people in central Ethiopia. The Gurage languages are written using the Ge'ez script, which is also used for other Ethiopian languages. The Gurage languages are not always mutually intelligible. 

\textbf{\textit{Tigrinya}} is spoken by about 9 million people, primarily in Eritrea and Ethiopia. It is written in the Ge'ez script, which is also used for Amharic, but the grammar and usage of Tigrinya differs significantly from Amharic. 
\subsubsection{Nilo-Saharan language family}
Nilo-Saharan languages are a group of languages that form one of the four language families on the African continent \cite{dimmendaal2019linguistic}. The family covers major areas east and north of Lake Victoria in East Africa and extends westward to the Niger Valley in Mali, West Africa \cite{comrie2002languages}. Nilo-Saharan constitutes ten distinct and separate language families, including Eastern Sudanic.

\noindent\textbf{Eastern Sudanic languages} are a group of ten families of languages that constitute a branch of the Nilo-Saharan language family. Eastern Sudanic languages are spoken from southern Egypt to northern Tanzania. The languages used in our study by this group are given below.  

\textbf{\textit{Majang}} is spoken by the Majangir people of Ethiopia. It is a member of the Surmic language cluster, but it is the most isolated one in the group. It is classified as part of the Eastern Sudanic branch of the Nilo-Saharan language family. 
The Majang people live in scattered settlements in southwestern Ethiopia. They live around the urban areas of Tepi and Mett'i, southwest of Mizan Teferi and towards Gambela. 

\textbf{\textit{Murle }} is spoken by the Murle people in South Sudan and Ethiopia. The language is also known as Ajibba, Beir, Merule, Mourle, and Murule. 
 The Murle language is part of the Surmic language family and has three dialects: Lotilla, Boma, and Olam. 
 The Murle people number between 300,000 and 400,000. They live in Pibor County in the southeastern Upper Nile (Jonglei)\\
\textbf{\textit{Nuer}} or Thok Naath is a West Nilotic language spoken by the Nuer people of South Sudan and western Ethiopia. The language is written in a Latin-based alphabet, similar to Dinka and Atuot. Over 900,000 people speak the Nuer language in diaspora communities in East Africa, Australia, and the USA. 


\section{Dataset}
\subsection{Dataset Collection}

We collected datasets for 16 languages from religious domains from a website\footnote{https://www.bible.com/}. In addition to that, for Amharic, Afaan Oromo, Somali, and Tigrinya, we collected publicly available datasets \cite{abate2019english,lakew2020low,vegi2022webcrawl} from different domains to create one benchmark dataset per language. For Dawuro, Gamo, Gofa, and Wolaita languages, we used \citet{tonja2021parallel} dataset to create benchmark results for fine-tuned models. A web crawler was used for each article to extract the Bible data from websites after identifying the structure of web documents. Python libraries such as requests, regular expression (RE), and Beautiful Soup (BS) were utilized to analyze website structure and extract article content from a given URL.
\subsection{Sentence Alignment}
After collecting the corpus for the languages, we aligned each sentence of the Ethiopian languages to a sentence in English data to prepare the dataset for the MT experiment. We followed the same procedure as \citet{tonja2023parallel} to perform sentence alignment. 
\subsection{Dataset Pre-processing}
After aligning the texts of the Ethiopian languages with their equivalent translations in English, we pre-processed the corpus
before splitting it for our experiments. The pre-processing steps included removing the numeric and special character symbols, etc. We also removed parallel sentences that contain less than five words. For the baseline experiments, we split the pre-processed corpus into training, development, and test sets in the ratio of 70:10:20, respectively. Table \ref{tab:language} shows detailed information on selected languages, language families, domain, and their dataset size.

\section{Baseline Models}
We used the following two approaches to evaluate the newly collected corpus's usability and our new benchmark dataset of four (amh, orm, som, and tir)  Ethiopian languages. 

\textbf{The baseline transformer} is a type of neural network architecture first introduced in the paper \textit{Attention Is All You Need} \cite{vaswani2017attention}. 
The key innovation of the Transformer architecture is the attention mechanism, which allows the network to selectively focus on different parts of the input sequence when making predictions. This contrasts traditional recurrent neural networks (RNNs), which process input sequentially and are prone to the vanishing gradient problem.

In the transformer architecture, multiple self-attention layers and feed-forward neural networks process elements of the input sequence in parallel. Each layer can be considered a "block" that takes the previous layer's output as input and applies its transformations to it. The self-attention mechanism allows the network to weigh the importance of each element in the input sequence when making predictions. In contrast, the feed-forward networks help to capture non-linear relationships among the components.

Transformers are state-of-the-art approaches widely used in NLP tasks such as MT, text summarization, and sentiment analysis. Table \ref{tab:my-tr} shows parameters set up for the transformer model.
\begin{table}[!ht]

\begin{tabular}{ll}
 \hline
\textbf{Parameters} & \textbf{Values} \\ \hline
encoder\_layer &6 \\
encoder\_attention\_head & 4 \\
decoder\_layer & 6 \\
batch\_size & 512  \\
batch\_type & token \\
decoder\_attention\_head & 8 \\
hidden\_size      & 256             \\ 
embed\_dim    & 256             \\ 
dropout            & 0.2             \\ 
beam\_size & 5 \\
optimizer          & adam            \\ 
tokenizer\_type & sentencepiece \\
max\_input\_length & 150 \\\hline
\end{tabular}
\caption{Parameters used for transformer training}
\label{tab:my-tr}
\end{table}

\textbf{Fine tuning} is the process of using a pre-trained MT model and adapting it to a specific translation task, such as translating between a particular language pair or in a specific domain. The process of fine-tuning involves taking the pre-trained model, which has already learned representations of words and phrases from a large corpus of text, and training it on a smaller dataset of specific task examples. This involves updating the pre-trained model's parameters to better capture the patterns and structures in the target translation task.
   
Fine-tuning can be helpful in MT because it allows the pre-trained model to quickly adapt to a new task without having to train a new model from scratch. This is especially beneficial when working with limited data or when there is a need to quickly adapt to changing translation requirements. 
We used \textbf{M2M100-48} a multilingual encoder-decoder (seq-to-seq) model trained for many-to-many multilingual translation \cite{fan2021beyond}. We used a model with 48M parameters due to computing resource limitations.
We used the following parameters to fine-tune the m2m100 model.
\begin{table}[!ht]
\begin{tabular}{ll}
 \hline
\textbf{Parameters} & \textbf{Values} \\ \hline
encoder\_layer &12 \\
encoder\_attention\_head & 16 \\
decoder\_layer & 12 \\
batch\_size & 512  \\
batch\_type & token \\
decoder\_attention\_head & 16 \\
hidden\_size      & 4096             \\ 
embed\_dim    &  1024             \\ 
attention\_dropout            & 0.1             \\ 
beam\_size & 5 \\ \hline
\end{tabular}
\caption{Parameters used for m2m100-48 fine-tuning}
\label{tab:my-tr}
\end{table}

\begin{figure*}
    \centering
    \includegraphics[scale=0.3]{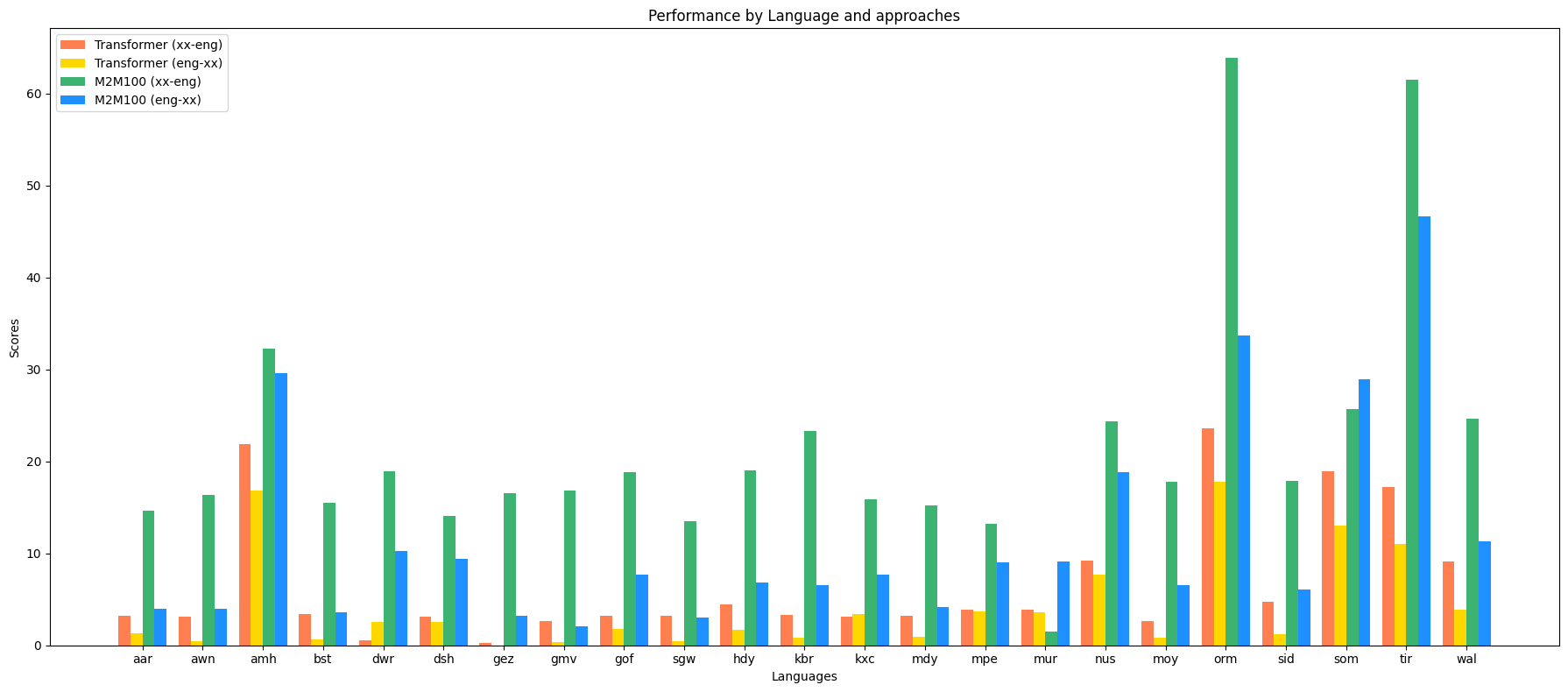}
    \caption{Benchmark translation results for transformer and fine-tuned approaches in both (from and to English/Ethiopian languages) direction }
    \label{fig:enter-label}
\end{figure*}

\begin{table*}[!ht]
\centering
  \resizebox{\textwidth}{!}{
\begin{tabular}{c|lllllllllllllllllllllll|l}\hline
\multirow{2}{*}{Model} & \multicolumn{24}{c}{en-xx} \\
                       & \textbf{aar} & \textbf{awn} & \textbf{amh} & \textbf{bst} & \textbf{dwr} & \textbf{dsh} & \textbf{gez} & \textbf{gmv} & \textbf{gof} & \textbf{sgw} & \textbf{hdy} & \textbf{kbr} & \textbf{kxc} & \textbf{mdy} & \textbf{mpe} & \textbf{mur} & \textbf{nus} & \textbf{moy} & \textbf{orm} & \textbf{sid} & \textbf{som} & \textbf{tir} & \textbf{wal} &\textbf{Avg.}\\\hline
                       & \multicolumn{24}{c}{\textbf{Bleu Score}}                                                                                                \\\hline
Transformer  & 1.28    & 0.41    & 16.79    &0.6     & 2.57    &2.51     &0.01     &0.34     &1.82     & 0.41    &1.69     &0.87     & 3.36    & 0.90    & 3.65    & 3.58    &7.73     & 0.87 &17.8& 1.19    & 13.06    &11.07    &3.84   & 4.18 \\\hline
m2m100-fine-tuned  & 3.95    & 3.93    &29.63     & 3.61    &10.23     & 9.45    &3.25     & 2.03    & 7.65    &3.04     &6.80     & 6.58    &7.69     & 4.15    &9.03     & 9.10    & 18.79    &6.58 &33.7  &6.10     &   28.9 & 46.63    &11.32 &\underline{11.83}   \\ \hline
\end{tabular}
}
\caption{Benchmark translation results from English to Ethiopian languages}
\label{tab:from_eng}
\end{table*}

\begin{table*}[!ht]
  \resizebox{\textwidth}{!}{
\begin{tabular}{c|lllllllllllllllllllllll|l}\hline
\multirow{2}{*}{Model} & \multicolumn{24}{c}{xx-en} \\
                       & \textbf{aar} & \textbf{awn} & \textbf{amh} & \textbf{bst} & \textbf{dwr} & \textbf{dsh} & \textbf{gez} & \textbf{gmv} & \textbf{gof} & \textbf{sgw} & \textbf{hdy} & \textbf{kbr} & \textbf{kxc} & \textbf{mdy} & \textbf{mpe} & \textbf{mur} & \textbf{nus} & \textbf{moy} & \textbf{orm} & \textbf{sid} & \textbf{som} & \textbf{tir} & \textbf{wal} &\textbf{Avg.}\\\hline
                       & \multicolumn{24}{c}{\textbf{Bleu Score}}                                                                                                \\\hline
Transformer            &  3.18   &  3.14   & 21.9    &   3.39  &  0.52   &  3.07   &  0.28   & 2.68    & 3.21    & 3.18    & 4.42    &  3.26   &  3.14   &  3.21   &  3.91   &  3.92   & 9.23    &  2.63   &23.6     & 4.77    &18.9     &17.2     & 9.16   & 6.60 \\\hline
m2m100-fine-tuned  & 15.61    & 16.32    & 65.34    & 15.47    & 18.92    & 14.11    & 16.57    &16.79     &18.79     &13.52     & 19.04    & 23.27    & 15.90    & 15.20    &  13.26   &1.48     &24.40     &17.78     & 63.9    & 17.86    &25.71     & 61.50    & 24.62 &\underline{21.79} \\ \hline
\end{tabular}
}
\caption{Benchmark translation results from Ethiopian languages to English}
\label{tab:to_eng}
\end{table*}

\section{Results and Discussions}
We evaluated the above approaches in bidirectional translation from Ethiopian languages to English and From English to Ethiopian languages. We used Sacrebleu \cite{post-2018-call} evaluation metrics to evaluate translation models.
Tables \ref{tab:from_eng} and \ref{tab:to_eng} show the translation results in both directions.
\subsection{Using English as a source language}
Table \ref{tab:from_eng} shows the translation results from English to Ethiopian languages. When comparing the results of the two approaches, we observe poor performance when using a transformer rather than fine-tuning the m2m100 model. As we can see from the result, the performance of the transformer model also varies in the ranges of 0.01 -- 17.8 spBLEU from language to language with different corpus sizes. This shows that a bilingual translation model trained from scratch performs poorly for low-resource language training compared to other approaches like fine-tuning multilingual translation models. Fine-tuning the multilingual model shows better results than the model built from scratch for English to Ethiopian language translation. In the fine-tuning approach, we can also observe a clear score difference between languages with larger corpora (amh, orm, tir, som) and others (e.g awn, aar, bst, etc.). This shows that fine-tuning the multilingual model will work well for languages with the largest (e.g. orm, amh) corpus sizes than languages with small (e.g. awn, bst, etc.) corpus sizes. We can also see from the results that both approaches work well for languages with mixed-domain texts compared to one domain (religion).

\subsection{Using English as a target language}
Table \ref{tab:to_eng} shows the translation result when using English as a target language. Similarly, as we can see from the results, the transformer model performs poorly compared to the fine-tuned model when translating from Ethiopian languages to English. Compared to Table \ref{tab:from_eng}, translating to English shows improvements in the transformer model for similar languages. We observe that the fine-tuned model shows better Bleu scores when translating to English than when translating to Ethiopian languages. The results show that languages with large datasets have the highest performance. This shows that both models show improvements when translating from Ethiopian to English, while when translating from English to Ethiopian languages, the model is struggling with translation.

\section{Conclusion and Future Works}
This paper presents EthioMT, a new MT corpus for low-resource Ethiopian languages paired with English, and discusses MT experiments with results. We also present a new benchmark dataset for four Ethiopian languages collected from public repositories. We obtained benchmark results with new train, validation, and test set splits and evaluated the new corpus and new benchmark dataset using a transformer and fine-tuning multilingual translation models. From the two approaches, fine-tuning of the multilingual model outperformed the transformer approach in both translation directions.

In the future, we will work to increase the corpus sizes of the low-resource languages by extracting text from scanned documents and different sources. In addition, we will evaluate different MT approaches to low-resource languages to improve performance.   

\nocite{*}
\section{Bibliographical References}\label{sec:reference}
\bibliographystyle{lrec-coling2024-natbib}
\bibliography{lrec-coling2024-example}


\end{document}